\documentclass{sig-alternate}

\usepackage[english]{babel}
\usepackage{latexsym}
\usepackage{bm}
\usepackage{graphicx}
\usepackage{booktabs}
\usepackage{multirow}
\usepackage{amssymb, amsmath}
\usepackage{array}
\usepackage{mdwlist}
\usepackage{subcaption}
\usepackage{diagbox}
\usepackage[hidelinks]{hyperref}

\begin{document}
%

\title{How to Generate a Good Word Embedding?}
%
%
%
%
%

\numberofauthors{1} 
%
\author{
%
%
\alignauthor Siwei Lai, Kang Liu, Liheng Xu, Jun Zhao\\
       \affaddr{National Laboratory of Pattern Recognition (NLPR)}\\
       \affaddr{Institute of Automation, Chinese Academy of Sciences, China}\\
       \email{\{swlai, kliu, lhxu, jzhao\}@nlpr.ia.ac.cn}
}
\date{20 July 2015}
\maketitle
\begin{abstract}

We analyze three critical components of word embedding training: the model, the corpus, and the training parameters.
We systematize existing neural-network-based word embedding algorithms and compare them using the same corpus.
We evaluate each word embedding in three ways: analyzing its semantic properties, using it as a feature for supervised tasks and using it to initialize neural networks.
We also provide several simple guidelines for training word embeddings. 
First, we discover that corpus domain is more important than corpus size. We recommend choosing a corpus in a suitable domain for the desired task, after that, using a larger corpus yields better results.
Second, we find that faster models provide sufficient performance in most cases, and more complex models can be used if the training corpus is sufficiently large.
Third, the early stopping metric for iterating should rely on the development set of the desired task rather than the validation loss of training embedding.

\end{abstract}



\section{Introduction}
\label{sec:intro}
Word embedding \cite{Bengio:2003,Collobert:2008}, also known as distributed word representation \cite{turian:2010}, can capture both the semantic and syntactic information of words from a large unlabeled corpus \cite{Mikolov:2013NAACL} and has attracted considerable attention from many researchers.
In recent years, several models have been proposed, and they have yielded state-of-the-art results in many natural language processing (NLP) tasks.
Although these studies always claim that their models are better than previous models based on various evaluation criteria, there are still few works that offer fair comparisons among existing word embedding algorithms. In other words, we still do not know how to design an effective word embedding algorithm for a specific task. Therefore, this paper focuses on this issue and presents a detailed analysis of several important features for training word embeddings, including the model construction, the training corpus and the parameter design. To the best of our knowledge, no such study has previously been performed.

\begin{table}[t]
	\centering \small
	\begin{tabular}{c|c|c}
		\toprule
		         \textbf{Model}           & \textbf{Relation of $w,c$} & \textbf{Representation of $c$} \\ \midrule
		Skip-gram \cite{Mikolov:2013ICLR} &      $c$ predicts $w$      &           one of $c$           \\
		  CBOW \cite{Mikolov:2013ICLR}    &      $c$ predicts $w$      &            average             \\
		              Order               &      $c$ predicts $w$      &         concatenation          \\
		      LBL \cite{Mnih:2007}        &      $c$ predicts $w$      &        compositionality        \\
		     NNLM \cite{Bengio:2003}      &      $c$ predicts $w$      &        compositionality        \\
		   C\&W \cite{Collobert:2008}     &        scores $w,c$        &        compositionality        \\ \bottomrule
	\end{tabular}
	
	\caption{A summary of the investigated models, including how they model the relationship between the target word $w$ and its context $c$, and how the models use the embeddings of the context words to represent the context.}
	\label{tab:models}
\end{table}

To design an effective word embedding algorithm, we must first clarify the model construction. We observe that almost all methods for training word embeddings are based on the same \emph{distributional hypothesis}: words that occur in similar contexts tend to have similar meanings \cite{Harris:1954}.
Given this hypothesis, different methods model the relationship between a word $w$ (the target word) and its context $c$ in the corpus in different ways, where $w$ and $c$ are embedded into vectors.
Generally, existing methods vary in two major aspects in model construction: (i) the relationship between the target word and its context and (ii) the representation of the context. Table~\ref{tab:models} shows a brief comparison of commonly used models. 
In terms of aspect (i), the first five models are identical. They use an object similar to the conditional probability $p(w|c)$, which predicts the target word $w$ based on its context $c$. C\&W \cite{Collobert:2008} uses an object similar to the joint probability, where the $w$, $c$ pairs in the corpus are trained to obtain higher scores. 
In terms of aspect (ii), the models use four different types of methods. They are listed in ascending order of complexity from top to bottom in Table~\ref{tab:models}. 
Skip-gram \cite{Mikolov:2013ICLR} use the simplest strategy, which is to choose one of the words from the window of the target word and utilize its embedding as the representation of the context. CBOW \cite{Mikolov:2013ICLR} uses the average embedding of the context words as the context representation. Both of these methods ignore the word order to accelerate the training process.
However, Landauer \cite{landauer:2002computational} estimates that 20\% of the meaning of a text comes from the word order, whereas the remaining meaning comes from word choice. Therefore, these two models may lose some critical information. In contrast, the Order model (Section~\ref{sec:order}) uses the concatenation of the context words' embeddings, which maintains the word order information. Furthermore, the LBL \cite{Mnih:2007}, NNLM \cite{Bengio:2003} and C\&W models add a hidden layer to the Order model. Thus, these models use the semantic compositionality \cite{hermann2014phd} of the context words as the context representation.
Based on the above analysis, we wish to know \emph{which model performs best?} and \emph{what selection we should make in terms of the relationship between the target word and its context as well as the different types of context representation?}

Moreover, the ability to train precise word embeddings is strongly related to the training corpus. Different training corpora of different sizes and different domains can considerably influence the final results. Thus, we additionally wish to know \emph{how the corpus size and corpus domain affect the performance of a word embedding?}

Finally, training precise word embedding strongly depends on certain parameters, such as the number of iterations and the dimensionality of the embedding. Therefore, we attempt to analyze the choice of these parameters. In particular, we wish to know \emph{how many iterations should be applied to obtain a sufficient word embedding while avoiding overfitting?} and \emph{what dimensionality we should choose to obtain a sufficiently good embedding?}

To answer these questions objectively, we evaluate various word embedding models on eight total tasks of three major types. We argue that all current applications of word embeddings are covered by these three types. The first type concerns calculating the word embedding's semantic properties. We use the standard WordSim353 dataset \cite{finkelstein:2002} and the classic TOEFL set \cite{landauer:1997solution} to check whether the distances of the embeddings in a vector space are consistent with human intuition. 
The second type concerns employing word embeddings as features for existing NLP tasks. Specifically, we use text classification and named entity recognition as the tasks for evaluation.
The third type concerns employing word embedding to initialize neural network models. 
Specifically, we employ a sentiment classification task based on a convolutional neural network and a part-of-speech tagging task using the neural network proposed by Collobert et al.~\cite{Collobert:2011}. Through different tasks with different usage types of word embeddings, we attempt to determine which type of model design is more suitable for each particular task (the first question). Moreover, we vary the corpus size and domains to attempt to answer the second question.

This paper has two main contributions: First, we systematize existing neural word embedding algorithms from the perspective of the relationship between the target word and its context, and the representation of the context (Section~\ref{sec:model})\footnote{Training and evaluation code are publicly available at: https://github.com/licstar/compare}. Second, we comprehensively evaluate the models on multiple tasks (Section~\ref{sec:eval}), and attempt to give some guidelines for generating a good word embedding:

\begin{itemize}
	\item For model construction, more complex models require a larger training corpus to outperform simpler models. Faster (simpler) models are sufficient in most cases. In semantic tasks, the models that predict the target word (the first five models in Table~\ref{tab:models}) demonstrate higher performance than the model that scores the combination of the target word and its context (the C\&W model). (Section~\ref{sec:expr_model})
	\item When choosing a corpus in a suitable domain, using larger corpus is better. Training on a large corpus generally improves the quality of word embeddings, and training on an in-domain corpus can significantly improve the quality of word embeddings for a specific task. More importantly, we discover that corpus domain is more important than corpus size. (Section~\ref{sec:expr_corpus})
	\item The validation loss is not a good early stopping metric for training word embeddings; instead, the best approach is to check the performance of the development set for that task. If evaluating the task is time consuming, the performance for other tasks can be used as an alternative. (Section~\ref{sec:expr_iter})
	\item For the tasks that analyzing the semantic properties of a word embedding, larger dimensions can provide a better performance. For NLP tasks that utilize an embedding as a feature or for initialization, a dimensionality of 50 is sufficient. (Section~\ref{sec:expr_dim})
\end{itemize}

\section{Models}
\label{sec:model}
In this section, we describe and analyze the word embedding models to be compared in the experiments.
We denote the embedding of word $w$ by $\bm{e}(w)$.

\subsection{Model Overview}

\subsubsection{NNLM}

\begin{figure}[th]
	\centering
	\includegraphics[width=0.47\textwidth]{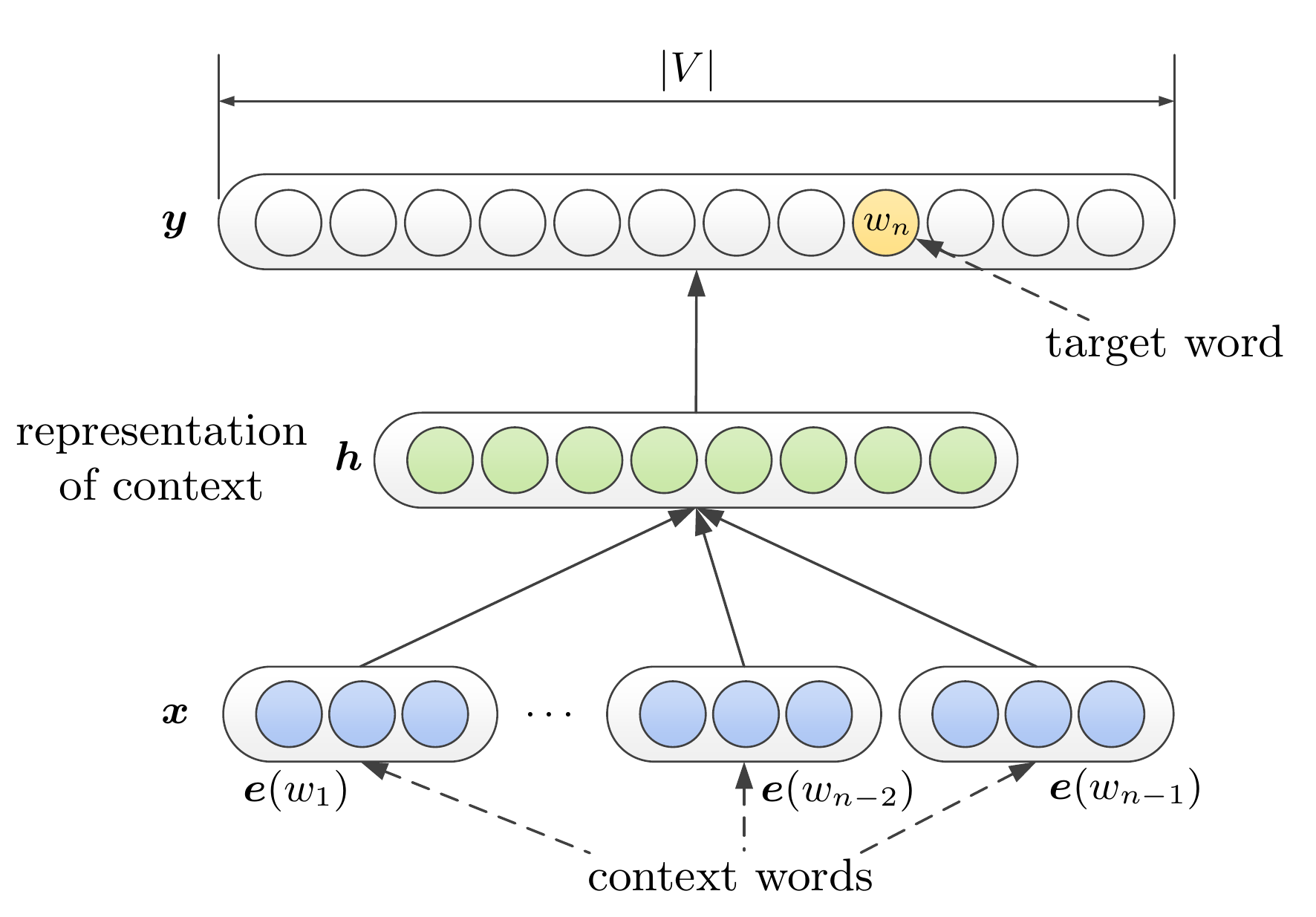}
	\caption{The model structure of the NNLM.}
	\label{fig:nnlm}
\end{figure}
Bengio et al.~\cite{Bengio:2003} first proposed a Neural Network Language Model (NNLM) that simultaneously learns a word embedding and a language model.
A language model utilizes several previous words to predict the distribution of the next word. For each sample in the corpus, we maximize the log-likelihood of the probability of the last word given the previous words.
For example, for a sequence $w_1, w_2, \dots, w_n$ in a corpus, we need to maximize the log-likelihood of
\begin{equation}
	p(w_n|w_1, \dots, w_{n-1})
\end{equation}
where we refer to the word we need to predict ($w_n$) as the target word. This model uses a concatenation of the previous words' embeddings as the input:
\begin{equation}
	\bm{x} = [\bm{e}(w_{1}), \dots, \bm{e}(w_{n-2}), \bm{e}(w_{n-1})]
\label{f:nnlm_x}
\end{equation}
The model structure is a feed-forward neural network with one hidden layer:
\begin{equation}
	\bm{h} = \tanh(\bm{d}+H\bm{x})
\label{f:nnlm_h}
\end{equation}
\begin{equation}
	\bm{y} = \bm{b} + U\bm{h}
\label{f:nnlm_h_y}
\end{equation}
where $U$ is a transformation matrix, $\bm{b}$ and $\bm{d}$ are bias vectors. The final step is to apply $\bm{y}$ to a softmax layer to obtain the probability of the target word.

\subsubsection{LBL}
The Log-Bilinear Language Model (LBL) proposed by Mnih and Hinton \cite{Mnih:2007} is similar to the NNLM. The LBL model uses a log-bilinear energy function that is almost equal to that of the NNLM and removes the non-linear activation function $\tanh$.
Mnih and Hinton \cite{mnih:2009scalable} accelerated this model by using the hierarchical softmax \cite{morin:2005hierarchical} to exponentially reduce the computational complexity; the resulting method is named the HLBL.
Mnih and Kavukcuoglu \cite{mnih:2013learning} further accelerated the method by developing the ivLBL model, which is very similar to the Skip-gram model. This model uses noise-contrastive estimation (NCE) \cite{gutmann:2012noise} to estimate the conditional probability of the target word.

\subsubsection{C\&W}
\begin{figure}[th]
	\centering
	\includegraphics[width=0.47\textwidth]{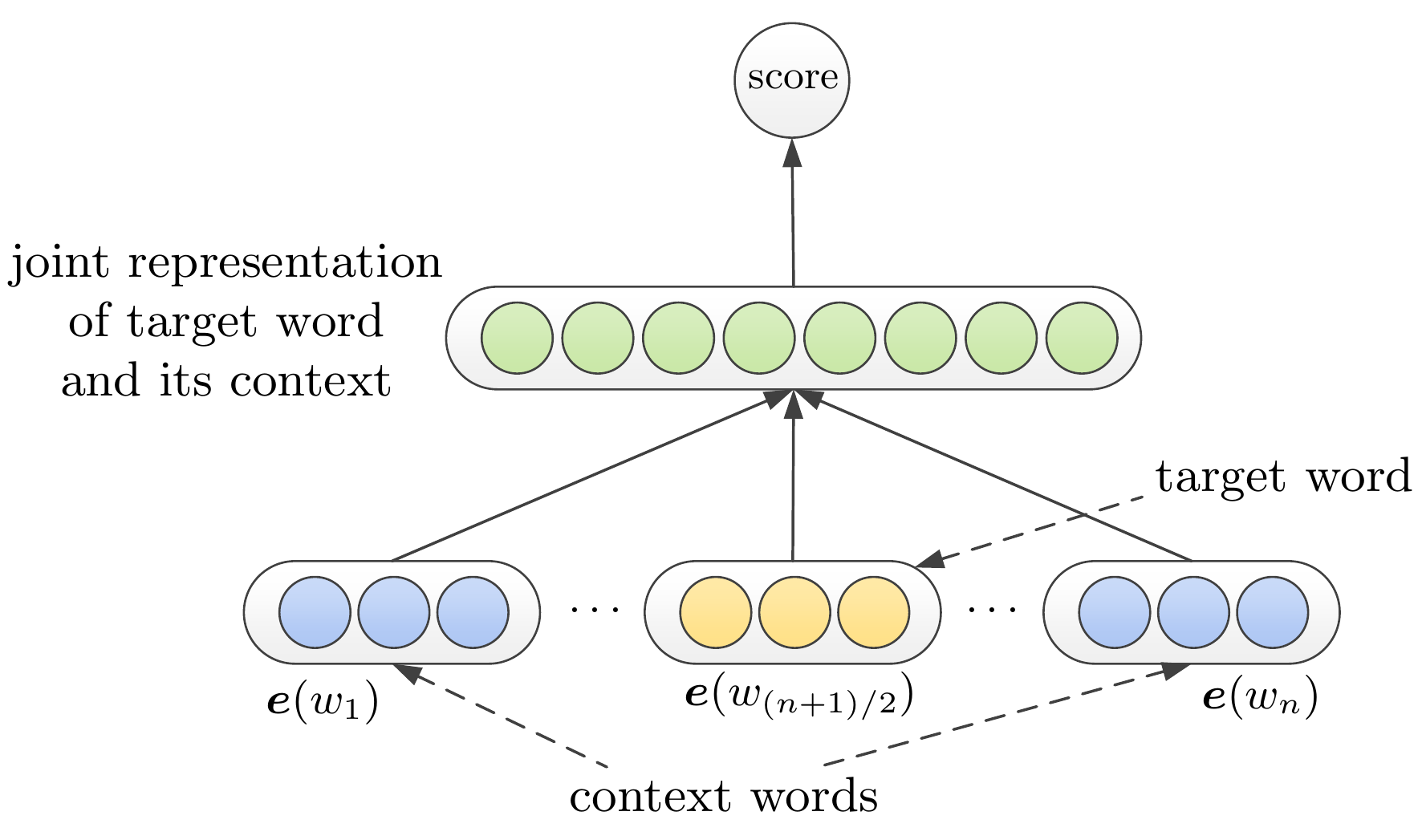}
	\caption{The structure of the C\&W model.}
	\label{fig:cw}
\end{figure}

Collobert and Weston (C\&W) \cite{Collobert:2008} proposed a model that trains only the word embedding. The C\&W model does not predict the target word; therefore, it combines the target word and its context and then scores them. 
The target word in this model is the central word in a sequence. The input is the concatenation of the target word's embedding and the context words' embeddings, i.e., the concatenation of the sequence $\bm{e}(w_1), \dots, \bm{e}(w_n)$.
The scoring function is a one-hidden-layer neural network.
The training target is to maximize the score of a corpus sequence while minimizing the score of a noise sequence pairwisely, or formally, to minimize
\begin{equation}
	\max(0, 1-s(w,c)+ s(w',c))
\end{equation}
In the noise sequence, the target word $w$ is replaced with a random word $w'$ from the vocabulary.

\subsubsection{CBOW and Skip-gram}
The CBOW and Skip-gram models \cite{Mikolov:2013ICLR} attempt to minimize computational complexity.
The CBOW model uses the average embedding of the context words as the context representation. The Skip-gram model uses one of the context words as the representation of the context. Both models neglect word order information. (In fact, the weighting strategy for context words maintains a small amount of word order information.) Only logistic regression is applied to the context representation to predict the target word.
Mikolov et al.~\cite{Mikolov:2013NIPS} further improved these models by introducing negative sampling as a simplification of the softmax approach based on the concepts applied in the two previously discussed models (ivLBL and C\&W). Negative sampling yields an improvement compared with the hierarchical softmax or NCE.

\subsubsection{Virtual Model: Order}
\label{sec:order}
The CBOW and Skip-gram approaches simplify the NNLM and LBL models by removing the word order information and the hidden layer.
To analyze the use of word order information, we introduce a virtual model ``Order'', whose complexity is between those of the CBOW and LBL models. This model maintains word order while removing the hidden layer.
In contrast to the CBOW model, this model uses the concatenation of the context words as input. In contrast to the LBL model, our model uses logistic regression instead of the log-bilinear structure.

\subsubsection{GloVe}
In addition to the neural network approaches, another line of research regarding word embedding is based on the word-context matrix \cite{turney:2010frequency,lebret:2014word}, in which each row corresponds to a word and each column corresponds to a context. The element in the matrix is relevant to the co-occurrence times of the corresponding word and context. 
These models are called count-based models \cite{baroni:2014don}.
The latest research on such matrix approaches is the Global Vectors model (GloVe)~\cite{Pennington:2014}, in which only the reconstruction errors of the non-zero values are minimized.

\subsection{Model Analysis}
\subsubsection{Relation between Target Word and its Context}
Existing neural word embedding models design the relation between the target word and the context in two different ways. Most models use the context to predict the target word. In NNLM, as shown in Figure~\ref{fig:nnlm}, the hidden layer $\bm{h}$ is the representation of the context. The dimension of the transformation matrix $U$ is $|V| \times |\bm{h}|$, in which $|V|$ is the size of the vocabulary, $|\bm{h}|$ is the dimension of the hidden layer. Each column in matrix $U$ can be regarded as an supplementary embedding of the corresponding word. We denote the supplementary embedding for word $w$ as $\bm{e}'(w)$. So, in these models, each word in vocabulary has two embeddings, $\bm{e}(w)$ when $w$ is in context, and $\bm{e}'(w)$ when $w$ is the target word. The energy function for word $w$ is $\bm{e}'(w)^\mathrm{T} \bm{h}$.
Contrast to the language model based approaches, C\&W model put the target word in the input layer and only maintain one embedding for each word. The energy function for word $w$ is $A\bm{e}(w)+ B\bm{c}$, in which $A$ and $B$ are transformation matrices, $\bm{c}$ is the representation of the context.
Therefore, the two types of models are considerably different.

\subsubsection{Representation of Context}
The models that predict the target word use different strategies to represent the context. Table~\ref{tab:models_context} shows the formalized representation for each model. The Skip-gram model choose one of the words in context and utilize its embedding as the representation. CBOW use average embedding of the context words, which contains the entire information of the context words. Order use the concatenation of the context words' embedding, which maintains the word order information. LBL and NNLM further add a hidden layer, which contains the compositionality of context words. NNLM has a non-linear transformation in addition to the LBL model, which makes the model more expressive.

\begin{table}[t]
	\centering
	\begin{tabular}{c|c}
		\toprule
		\textbf{Model} & \textbf{representation of the context} \\ \midrule
		Skip-gram    &           $\bm{e}(w_{i}), 1 \le i \le n-1$          \\
		CBOW      &            $\frac{1}{n-1}(\bm{e}(w_{1})+ \dots+ \bm{e}(w_{n-2})+ \bm{e}(w_{n-1}))$             \\
		Order      &         $[\bm{e}(w_{1}), \dots, \bm{e}(w_{n-2}), \bm{e}(w_{n-1})]$          \\
		LBL       &        $H[\bm{e}(w_{1}), \dots, \bm{e}(w_{n-2}), \bm{e}(w_{n-1})]$        \\
		NNLM      &        $\tanh(\bm{d}+H[\bm{e}(w_{1}), \dots, \bm{e}(w_{n-2}), \bm{e}(w_{n-1})])$        \\ \bottomrule 
	\end{tabular}
	\caption{The formalized representation of the context for the models that predict the target word. Assuming the target word is $w_n$, and its context words are $w_1, \dots, w_{n-1}.$}
	\label{tab:models_context}
\end{table}

\section{Tasks}
\label{sec:eval}
We evaluate the various word embedding models in eight tasks grouped into three major types. We argue that all current applications of word embeddings are covered by these three types.

\subsection{Embedding's Semantic Properties}
Word embedding models are designed based on the distributional hypothesis; therefore, words with similar meanings tend to have similar word embeddings. Baroni et al.~\cite{baroni:2014don} have argued that word embeddings are characterized by various semantic properties. We include several classic tasks in our evaluation. 

\textbf{ws}~~~
The WordSim353 set \cite{finkelstein:2002} contains 353 word pairs. It was constructed by asking human subjects to rate the degree of semantic similarity or relatedness between two words on a numerical scale.
The performance is measured by the Pearson correlation of the two word embeddings' cosine distance and the average score given by the participants.

\textbf{tfl}~~~
 The TOEFL set \cite{landauer:1997solution} contains 80 multiple-choice synonym questions, each with 4 candidates. For example, the question word \emph{levied} has choices: \emph{imposed} (correct), \emph{believed}, \emph{requested} and \emph{correlated}. We choose the nearest neighbor of the question word from the candidates based on the cosine distance and use the accuracy to measure the performance.

\textbf{sem \& syn}~~~
The analogy task \cite{Mikolov:2013ICLR} has approximately 9K semantic and 10.5K syntactic analogy questions.
The question are similar to ``\emph{man} is to (\emph{woman}) as \emph{king} is to \emph{queen}'' or ``\emph{predict} is to (\emph{predicting}) as \emph{dance} is to \emph{dancing}''. Following the previous work, we use the nearest neighbor of $\overrightarrow{queen}-\overrightarrow{king}+\overrightarrow{man}$ in the vocabulary as the answer. Additionally, the accuracy is used to measure the performance.
This dataset is relatively large compared to the previous two sets; therefore, the results using this dataset are more stable than those using the previous two datasets.

\subsection{Embedding as Features}
Word embedding models capture useful information from unlabeled corpora. Many existing works have directly used word embeddings as features to improve the performance of certain tasks.
 We use two tasks; the first task is a text classification task in which the word embedding is the only feature. The second task is a named entity recognition task in which the word embedding is used as an additional feature in a near state-of-the-art system.

\textbf{avg}~~~
This task uses the weighted average of the word embeddings as the representation of the text and subsequently applies a logistic regression to perform text classification. The weight for each word is its term frequency. 
We use the IMDB dataset \cite{maas:2011learning}. This dataset contains three parts. The training and test set are used to train and test the text classification model. The unlabeled set is used to train the word embeddings.

\textbf{ner}~~~
Named entity recognition (NER) is typically treated as a sequence labeling problem. In this task, we utilize word embeddings as the additional feature of the near state-of-the-art NER system \cite{ratinov:2009design}. The experimental settings and code follow Turian et al.~\cite{turian:2010}.
The performance is evaluated with the F1 score on the test set of the CoNLL03 shared task dataset.

\subsection{Embedding as the Initialization of NNs}
Erhan et al.~\cite{Erhan:2010} have demonstrated that a better initialization can cause a neural network model to converge to a better local optimum. In recent neural network approaches to NLP tasks, word embeddings have been used to initialize the first layer.
In contrast to using word embeddings as features, using them for initialization results in the word embeddings being modified during the training of the neural network model.

\textbf{cnn}~~~
We use Convolutional neural networks (CNN) according to Kim \cite{kim:2014} to perform sentence-level sentiment classification on the Stanford Sentiment Treebank dataset \cite{socher:2013treebank}.
This dataset is relatively small; the initialization of other parameters in the CNN is sensitive to the result. We repeat our experiment five times with the same word embedding and different random initialized neural network parameters. 
For each experiment, we use the accuracy of the test set  when the development set achieved the best performance. 
We report the average accuracy of the five experiments.

\textbf{pos}~~~
The part-of-speech (POS) tagging task is a sequence labeling problem. We use the neural network proposed by Collobert et al.~\cite{Collobert:2011} to perform POS tagging on Wall Street Journal data \cite{toutanova:2003feature}.
The performance is measured by the accuracy on the test set when the development set achieved the best performance. 

\section{Experiments and Results}

In this section, to answer the questions posed in the Section~\ref{sec:intro}, we perform  comprehensive comparison among the considered models trained on different corpora with various parameters using the eight tasks in Section~\ref{sec:eval}. Table~\ref{tab:setting} shows the detailed experimental settings. 

\begin{table}[h!]
	\centering
	\begin{tabular}{c m {0.3\textwidth}} \toprule
		
		\textbf{Type} &  \textbf{Setting} \\
		\midrule
		Model & GloVe, Skip-gram, CBOW, Order,
		
		 LBL, NNLM, C\&W\\
		\midrule
		Corpus & Wiki: 100M, 1.6B;
		
		NYT: 100M, 1.2B;
		
		W\&N: 10M, 100M, 1B, 2.8B;
		
		IMDB: 13M;\\
		\midrule
		Para. & dimensionality: 10, 20, 50, 100, 200
		
		fixed window size: 5\\
		\bottomrule
	\end{tabular}
	\caption{A summary of the settings we used to investigate the training of word embeddings, including the chosen models, corpora, and hyperparameters.}
	\label{tab:setting}
\end{table}

\subsection{Performance Gain Ratio}
In order to compare the models across the eight tasks in Section~\ref{sec:eval}, we find it lack of a unified evaluation metric because of the following two reasons:

First, the performance metrics for the different tasks lie in different ranges. For example, the performances for the \textit{avg} task are always in the range of 70\% to 80\%, whereas the performances for the \textit{pos} task are typically greater than 96\% (refer to Table~\ref{tab:best_all_normal}). 

Second, the variances of the performances for the tasks are at different scales. For example, in the \textit{ws} task, the best model demonstrates a performance of 63.89\%, whereas the worst performance is merely 46.17\%. By contrast, in the \textit{avg} task, the variance among models is relatively small; the best performance is 74.94\%, and even the worst model achieves a performance of 73.26\%.

For the two reasons mentioned above, we may encounter some difficulty in comparing the word embeddings across tasks; in other words, it is not easy to determine whether a given embedding is generally useful for these tasks. Moreover, if, for a given task, one embedding performs better than another by one percentage point, it may also be unclear whether this difference truly represents a significant improvement or whether it can merely be attributed to the randomness inherent to the training task models.

To address these problems, we propose a new indicator, the \textit{Performance Gain Ratio} ($PGR$), in place of the original evaluation metrics (accuracy, F1, etc.).
The $PGR$ of embedding $a$ with respect to embedding $b$ is defined as
\begin{equation}
PGR (a,b)=\frac{p_a - p_{rand}}{p_b - p_{rand}} \times 100\%
\label{f:pgr}
\end{equation}
where $p_x$ is the performance of embedding $x$ for a given task and $p_{rand}$ is the performance achieved by a random embedding. Here, a random embedding is a word embedding of the same dimensionality as embedding $x$, where each dimension is a uniformly distributed random variable ranging from $-1$ to $1$. The numerator and denominator in Equation~(\ref{f:pgr}) are the \textit{Performance Gains} of embeddings $a$ and $b$, respectively.
The \textit{Performance Gain} of a word embedding $x$ refers to the increment in performance achieved by embedding $x$ compared with the performance achieved by a random embedding.
It is necessary to subtract the performance of a random embedding, because for the \textit{tfl} task, a random guess from among the four choices is expected to have 25\% accuracy, whereas for the tasks in which the embedding is used as a feature or for initialization, the model can achieve non-zero performance even without the aid of a word embedding.
Therefore, the performance increment of a given word embedding compared with a random embedding reflects the real performance gain offered by that embedding.

Embedding $b$ is chosen to be the best embedding for a given setting; thus, we call this measure simply the $PGR$ of embedding $a$. With this definition, the $PGR$ is a number smaller than $100\%$. If $PGR=100\%$, then the embedding achieved the best result among all embeddings for the same setting. If $PGR=0$, then the embedding may contain no more useful information than a random embedding. If $PGR < 0$, then the embedding is detrimental to the task.

In the way, we argue that the \textit{Performance Gain Ratio} normalizes the performances on all tasks to the same scale, thereby simplifying the analysis across multiple tasks. In the following experiments, we will use the $PGR$ when we need to compare embeddings across tasks.

\subsection{Model and Implementation}
\label{sec:expr_model}

In this section, we investigate how the different models perform in the eight evaluation tasks. In particular, we will compare the different types of context representations and the different relations between the target word and its context.

\subsubsection{Implementation Details}
To fairly compare the different models, we use the same implementation for all models.
Our GloVe implementation is based on the GloVe toolkit\footnote{http://nlp.stanford.edu/projects/glove/}. The CBOW and Skip-gram implementations are based on the \texttt{word2vec} toolkit\footnote{https://code.google.com/p/word2vec/}. The other models are modified from the CBOW implementation in \texttt{word2vec}. The Order model uses the concatenation of the context words instead of the average. The LBL model adds one linear hidden layer to the Order model. The NNLM adds the $\tanh$ activation function to the LBL model. The C\&W model moves the target word to the input of the neural network, leaving only one node in the softmax layer.
For all models, we use the center word in the window as the target word.
For the neural-network-based models (all models except GloVe), we use subsampling with $t=10^{-4}$ and negative sampling with 5 negative samples \cite{Mikolov:2013NIPS}.
We also make two modifications. First, the GloVe toolkit and \texttt{word2vec} use different weighting strategies for context words. We remove these weightings for both models. Levy et al.~\cite{levy:2015} demonstrated that these weightings have little influence to the task performances. Second, the original implementation of \texttt{word2vec} uses stochastic gradient descent with a decaying learning rate as the optimization algorithm. We modified this implementation using AdaGrad \cite{duchi:2011adaptive} to mirror the implementation of GloVe. The learning rate is set to 0.1, which causes the new optimization method to perform similarly to its performance using the original decaying learning rate.

\subsubsection{Comparison of Models}

\begin{table*}[ht]
	
		\centering
		\begin{tabular}{c|cccc|cc|cc}
			\toprule
			\textbf{Model} & \textbf{syn} & \textbf{sem} & \textbf{ws} & \textbf{tfl} & \textbf{avg} & \textbf{ner} & \textbf{cnn} & \textbf{pos}\\
			\midrule
			Random & 0.00 & 0.00 & 0.00 & 25.00 & 64.38 & 84.39 & 36.60 & 95.41\\
			\midrule
			GloVe & 40.00 & 27.92 & 56.47 & \textbf{77.50} & 74.51 & 88.19 & 43.29 & 96.42\\
			Skip-gram & 51.78 & \textbf{44.80} & \textbf{63.89} & 76.25 & \textbf{74.94} & \textbf{88.90} & 43.84 & 96.57\\
			CBOW & \textbf{55.83} & 44.43 & 62.21 & \textbf{77.50} & 74.68 & 88.47 & 43.75 & 96.63\\
			Order & 55.57 & 36.38 & 62.44 & \textbf{77.50} & \textbf{74.93} & 88.41 & \textbf{44.77} & \textbf{96.76}\\
			LBL & 45.74 & 29.12 & 57.86 & 75.00 & 74.32 & 88.69 & 43.98 & \textbf{96.77}\\
			NNLM & 41.41 & 23.51 & 59.25 & 71.25 & 73.70 & 88.36 & 44.40 & 96.73\\
			C\&W & 3.13 & 2.20 & 46.17 & 47.50 & 73.26 & 88.15 & 41.86 & 96.66\\
			\bottomrule
		\end{tabular}
			\caption{Best results for 50-dimensional embeddings train by each model on the W\&N corpus (2.8B tokens).}
		\label{tab:best_all_normal}

\end{table*}

For each model, we iterate until the model converges to or overfits all of the tasks.
Table~\ref{tab:best_all_normal} shows the best performance for each model trained on the W\&N corpus (a collection of Wikipedia and New York Times articles).
Overall, the best results are obtained using various models, from GloVe to LBL. Compared with random embedding, all investigated word embeddings demonstrate better performance on the tasks regardless of the model used.

To investigate the influence of \textbf{the relationship between the target word and its context}, we should compare the C\&W model, which scores the combination of the target word and its context, with the other models, which predict the target word.
The C\&W model exhibits lower performance compared with the other models in the semantic property tasks (\textit{syn}, \textit{sem}, \textit{ws}, and \textit{tfl}). Particularly in the analogy tasks (\textit{syn} and \textit{sem}), the results show that the C\&W model almost entirely lacks the feature of linear semantic subtraction.
To investigate the difference between the C\&W model and the models that predict the target word, we show the nearest neighbors of selected words in Table~\ref{tab:model_neighbor}.

\begin{table}[ht]
	\centering
	\begin{tabular}{c|ccc}
		\toprule
		\textbf{Model} & \textbf{Monday} & \textbf{commonly} & \textbf{reddish} \tabularnewline
		\midrule
		\multirow{5}{*}{CBOW} & Thursday & generically & greenish \tabularnewline
		& Friday & colloquially & reddish-brown \tabularnewline
		& Wednesday & popularly & yellowish\tabularnewline
		& Tuesday & variously & purplish \tabularnewline
		&  Saturday & Commonly & brownish \tabularnewline
		\midrule
		\multirow{5}{*}{C\&W} & 8:30 & often & purplish \tabularnewline
		& 12:50 & generally & pendulous \tabularnewline
		& 1PM & previously & brownish \tabularnewline
		& 4:15 & have & orange-brown \tabularnewline
		& mid-afternoon & are & grayish \tabularnewline
		\bottomrule
	\end{tabular}
	\caption{Nearest neighbors of selected words when trained using the CBOW and C\&W models on W\&N corpus.}
	\label{tab:model_neighbor}
\end{table}

From these cases, we find that the nearest neighbors of ``Monday'' trained by the CBOW model are the other days of the week. In contrast, when trained by the C\&W model, the nearest neighbors are times of day. The nearest neighbors of ``commonly'' also reveal a similar result: the CBOW model finds words that can replace the word ``commonly'', whereas the C\&W model will find words that are used together with ``commonly''. Most of the nearest neighbors of ``reddish'' in both models are colors. Except for the word ``pendulous'' in C\&W model, which can be used together with ``reddish'' to describe flowers.

Previous works have shown that models that predict the target word capture the paradigmatic relations between words \cite{rapp2002computation,sahlgren:2006word,levy:2014}. In short words, a target word links various groups of contexts that predict it, making them similar among themselves.
However, the C\&W model does not explicitly model these relations; thus, the relations are less paradigmatic compared with CBOW model. 
The results (Table~7) in original C\&W paper \cite{Collobert:2011} seems that the neighbors by C\&W model are paradigmatic relations. However, their embedding was generated by C\&W model and then fine tuned for four supervised tasks (POS tagging, NER, etc.). The supervised targets can also bring paradigmatic relations to word embeddings just like the target words in CBOW model.
We argue that the difference arises because the C\&W model places the target word in the input layer, and this structure may not capture paradigmatic relations as strongly compared with the models that predict the target word.

\begin{table}[ht]
	\centering
	\begin{tabular}{c|cccc}
		\toprule
		\textbf{Model} & 10M & 100M & 1B & 2.8B\\
		\midrule
		Skip-gram & 4+2 & 4+2 & 2+2 & 3+2\\
		CBOW & 1+1 & 3+3 & 4+1 & 4+1\\
		Order & 0+2 & 1+2 & 2+3 & 3+3\\
		LBL & 0+2 & 0+2 & 0+2 & 1+2\\
		NNLM & 0+2 & 0+3 & 0+3 & 0+2\\
		\bottomrule
	\end{tabular}
	\caption{The number of tasks for which a model ``wins'' on a certain corpus. In each cell, $a+b$ indicates that the model wins $a$ tasks of the first four tasks (semantic properties) and $b$ tasks of the last four tasks (used as a feature or for initialization).}
	\label{tab:model95}
\end{table}

To investigate the influence of \textbf{the representation of the context}, we further compare embeddings trained on corpora of different scales.
Table~\ref{tab:model95} reports the number of tasks for which each model achieves 95\% $PGR$ (for convenience, when this condition is satisfied, we say that the model ``wins''). A model that wins a task when trained on a 10M corpus should thus be compared with the best embedding trained on the 10M corpus.
We only focus on the listed five models because the only difference between these models are their representations of the context. According to the analysis in Section \ref{sec:intro}, we can view the different representations from the perspective of model complicity. Thus the experimental results are easy to explain. We note two observations based on Table~\ref{tab:model95}.

First, a simpler model can achieve better results on a small training corpus. The simplest model, Skip-gram, is the best choice when using a 10M- or 100M-token corpus. When using a larger corpus, more complex models, such as the Order model, win more often. For corpora larger than those we tested, the more complex models LBL and NNLM may be the best choices.

Second, for tasks in which the word embedding is used as a feature or for the initialization of neural networks, the choice of model does not significantly affect the results. In addition, based on the results given in Table~\ref{tab:best_all_normal}, the margins between the simple and complex models are relatively small. Therefore, simpler models are typically sufficient for real tasks.

We can now answer the first questions of \emph{which model performs best?} and \emph{what selection we should make in terms of the relationship between the target word and its context words as well as the different types of context representation?} In the case of a smaller corpus, a simpler model, such as Skip-gram, can achieve better results, whereas for a larger corpus, more complex models, such as CBOW and Order, are typically superior. Nevertheless, for real tasks, simpler models (Skip-gram, CBOW and Order) are typically adequate. In semantic tasks, the models that predict the target word demonstrate higher performance than does the C\&W model, which places the target word and the context together in the input layer.

\subsection{The Effect of the Training Corpus}
\label{sec:expr_corpus}
In this section, we investigate how the corpus size and corpus domain influence the performance of the word embedding.
We use two large-scale corpora and one small corpus with different domains in our experiments. These corpora are the Wikipedia (Wiki) dump\footnote{https://dumps.wikimedia.org/enwiki/}, the New York Times (NYT) corpus\footnote{https://catalog.ldc.upenn.edu/LDC2008T19} and the IMDB corpus. We combine the Wiki and NYT corpora to obtain the W\&N corpus. 
The vocabulary is set as the most frequent 200K words in both the Wiki and NYT corpora (each word should occur at least 23 times in both corpora). 
For each corpus, the words not in the vocabulary are ignored. We shuffle the corpus at the document level to decrease the bias caused by certain online learning models.
Small corpora are sampled uniformly from the larger corpus at the document level; thus, a 10M-token corpus is a subset of the corresponding 100M-token corpus.

We choose a representative model CBOW to analyze the influence of the corpus because of the space limitation. The other models yield similar results. Table~\ref{tab:corpus_cbow} shows the $PGR$ values for the embeddings trained on various corpora compared with the best results across all corpora. The best $PGR$ for each task is 100\%. 

\newcolumntype{L}[1]{>{\raggedright}p{#1}} 
\newcolumntype{R}[1]{>{\raggedleft}p{#1}} 
\newcolumntype{C}[1]{>{\centering}p{#1}}
\begin{table}[ht]
	\centering
	\begin{tabular}{l|R{0.43cm}R{0.43cm}R{0.43cm}R{0.43cm}R{0.43cm}R{0.43cm}R{0.43cm}R{0.43cm}}
		\toprule
		\textbf{Corpus} & \textbf{syn} & \textbf{sem} & \textbf{ws} & \textbf{tfl} & \textbf{avg} & \textbf{ner} & \textbf{cnn} & \textbf{pos}\tabularnewline
		\midrule
		NYT \scriptsize{1.2B}    & 93 & 52 & 90 & 98 & 50 & 76 & 85 & 96\tabularnewline
		~~~100M & 76 & 30 & 88 & 93 & 46 & 77 & 83 & 86\tabularnewline
		\midrule
		Wiki \scriptsize{1.6B}   & 92 & \scriptsize{\textbf{100}} & \scriptsize{\textbf{100}} & 93 & 51 & \scriptsize{\textbf{100}} & 86 & 94\tabularnewline
		~~~100M & 74 & 65 & 98 & 93 & 47 & 88 & 90 & 83\tabularnewline
		\midrule
		W\&N \scriptsize{2.8B}   & \scriptsize{\textbf{100}} & 89 & 95 & 93 & 50 & 97 & 91 & \scriptsize{\textbf{100}}\tabularnewline
		~~~1B   & 98 & 87 & 95 & \scriptsize{\textbf{100}} & 48 & 98 & 90 & 98\tabularnewline
		~~~100M & 79 & 63 & 97 & 96 & 51 & 85 & 92 & 86\tabularnewline
		~~~10M  & 29 & 27 & 76 & 60 & 42 & 49 & 77 & 42\tabularnewline
		\midrule
		IMDB \scriptsize{13M}   &32 & 21 & 55 & 82 & \scriptsize{\textbf{100}} & 26 & \scriptsize{\textbf{100}} & \mbox{-13}\tabularnewline
		\bottomrule
	\end{tabular}
	\caption{$PGR$ values for the CBOW model trained on different corpora.}
	\label{tab:corpus_cbow}
\end{table}

\subsubsection{Corpus Size}
From the results in Table~\ref{tab:corpus_cbow}, we can conclude that using a larger corpus can yield a better embedding, when the corpora are in the same domain.
We compare the full-sized NYT corpus with its 100M subset, the Wiki corpus with its subset, and the W\&N corpus with its three subsets. In almost all cases, the larger corpus is superior to the smaller corpus. The observed exceptions may be attributable to instability of the evaluation metrics.

Specially, in the \textit{syn} task (analogy task with syntax test cases, such as ``year:years law:\_\_''), the corpus size is the main driver of performance. The 100M-token subsets of the NYT, Wiki and W\&N corpora achieve similar results as does the 10M subset of the W\&N corpus and the 13M IMDB corpus. Clearly, different corpora use English in a similar manner, thereby producing similar syntax information.

\subsubsection{Corpus Domain}
In most of the tasks, the influence of the corpus domain is dominant. In different tasks, it impacts performance in the different ways:
\begin{itemize} 
	\item In the tasks involving the evaluation of semantic features, such as the analogy task with semantic test cases (\textit{sem}) and the semantic similarity task (\textit{ws}), the Wiki corpus is found to be superior to the NYT corpus. Even the 100M-token subset of Wiki corpus can achieve a better performance than the 1.2B-token NYT corpus.
	We believe that the Wikipedia corpus contains more comprehensive knowledge, which may benefit semantic tasks.
	\item The small IMDB corpus benefits the \textit{avg} and \textit{cnn} tasks, but it performs very poorly in the \textit{ner} and \textit{pos} tasks. The IMDB corpus consists of movie reviews from the IMDB website, which is the same source as that used for the training and test sets of the \textit{avg} and \textit{cnn} tasks. In-domain corpus is helpful for the tasks. Especially in the \textit{avg} task, the in-domain IMDB corpus achieves a $PGR$ almost twice than the second best one. On the other hand, these movie reviews are much more informal than the Wiki and NYT corpora, which is detrimental to the part-of-speech tagging (\textit{pos}) task.
\end{itemize}

\begin{table}[ht]
	\centering
	\begin{tabular}{c|ccc}
		\toprule
		\textbf{Corpus} & \textbf{movie} & \textbf{Sci-Fi} & \textbf{season}\tabularnewline
		\midrule
		\multirow{5}{*}{IMDB} & film & SciFi & episode\tabularnewline
		& this & sci-fi & seasons\tabularnewline
		& it & fi & installment\tabularnewline
		& thing & Sci & episodes\tabularnewline
		& miniseries & SF & series\tabularnewline
		\midrule
		\multirow{5}{*}{W\&N} & film & Nickelodeon & half-season\tabularnewline
		& big-budget & Cartoon & seasons\tabularnewline
		& movies & PBS & homestand\tabularnewline
		& live-action & SciFi & playoffs\tabularnewline
		& low-budget & TV & game\tabularnewline
		\bottomrule
	\end{tabular}
	\caption{Nearest neighbors of certain words when trained on the IMDB and W\&N corpora.}
	\label{tab:corpus_neighbor}
\end{table}

To intuitively illustrate how an in-domain corpus helps a given task, we present several selected words and their nearest neighbors in Table~\ref{tab:corpus_neighbor}. The neighbors of ``movie'' in the IMDB corpus consist of words such as ``this'', ``it'', and ``thing'', implying that the word ``movie'' can be treated as a stop word in the IMDB corpus. The neighbors of ``Sci-Fi'' in the IMDB corpus consist of other abbreviations for ``science fiction'', whereas the neighbors in the W\&N corpus are predominantly other genres. The word ``season'' in the IMDB corpus is predominantly associated with the episodes of a TV show, whereas it refers to the sports competition season in the W\&N corpus.
From these cases, we observe that an in-domain corpus improves performance because it allows more suitable word embeddings for the tasks to be obtained.

\subsubsection{Which is More Important, Size or Domain?}

\begin{table}[ht]
	\centering
	\begin{tabular}{c|rrrrr}
		\toprule
		\backslashbox{\scriptsize{W\&N}}{\scriptsize{IMDB}} & 20\% & 40\% & 60\% & 80\% & 100\% \\ \midrule
		                       +0\%                         &   91 &   94 &  100 &  100 &   100 \\
		                       +20\%                        &   79 &   87 &   91 &   96 &    99 \\
		                       +40\%                        &   68 &   86 &   88 &   92 &    98 \\
		                       +60\%                        &   65 &   79 &   85 &   88 &    93 \\
		                       +80\%                        &   64 &   75 &   84 &   87 &    92 \\
		                      +100\%                        &   64 &   70 &   83 &   86 &    88 \\ \bottomrule
	\end{tabular}
	\caption{$PGR$s of \textit{avg} task when trained on mixed corpora. Each column corresponds to the percentage of tokens we use in IMDB corpus, and each row corresponds to the percentage of tokens we use in the 13M-token subset of W\&N corpus. For example, the entry with value 68 in the third row and the first column means we combine 20\% of the IMDB corpus with 40\% of the 13M subset of W\&N corpus to train a word embedding, and obtain a $PGR$ of 68 on \textit{avg} task.}
	\label{tab:corpus_mix}
\end{table}

In previous sections, the experiment results demonstrated more in-domain training data would benefit to training a precise model. However, for a specific task, it may be impossible to obtain sufficient in-domain data. Increasing more training data may introduce more out-domain data and may contradict the demand of in-domain. In this case, should we keep the corpus pure or add the out-domain corpus? We further design an experiment to answer this question. In this experiment, we use the 13M-token IMDB corpus and a subset of W\&N corpus with 13M tokens. The subsets of the both corpora are mixed to train word embedding. Table~\ref{tab:corpus_mix} shows the $PGR$ of the \textit{avg} task for each embedding. For each column in the table, the elements are listed in ascending order of corpus size, while in descending order of the in-domain purity.
The experiment results demonstrate that no matter in which scale the IMDB corpus is, adding the W\&N corpus consistently decrease the performance. The pure IMDB corpus is better than the larger mixed corpus.

The answer to the question of \emph{how the corpus size and domain affect the word embedding?} is now clear. 
The corpus domain is more important than the corpus size. Using an in-domain corpus significantly improves the performance for a given task, whereas using a corpus in an unsuitable domain may decrease performance. For specific tasks, the use of a pure in-domain corpus yields better performance than does the use of a mixed-domain corpus. For corpora in the same domain, a larger corpus will allow better performance to be achieved.

\subsection{The Choice of the Training Parameters}

\subsubsection{Number of Iterations}
\label{sec:expr_iter}
The word embedding models presented in Section~\ref{sec:model} all iteratively optimize the objective function; thus, the performance of these word embeddings depends on the number of iterations.
If the iteration number is small, the word embeddings will exhibit a lack of learning. In contrast, if the iteration number is large, the model will be prone to overfitting.

In machine learning, early stopping is a form of regularization used to tackle with this problem.
The most widely used early stopping method is to stop the iterator when the loss on the validation set peaks \cite{prechelt1998early}. 
In word embedding training, the loss measures how well the model predicts the target word.
However, real tasks typically do not consist of predicting target words.
Therefore, the loss on the validation set is only a proxy for these real tasks, and in some cases, it may be inconsistent with the task performance. Therefore, it is worth investigating whether the traditional early stopping method is a good metric for stopping the training of word embeddings.


\begin{figure}[th]
	\centering
	\includegraphics[width=0.47\textwidth]{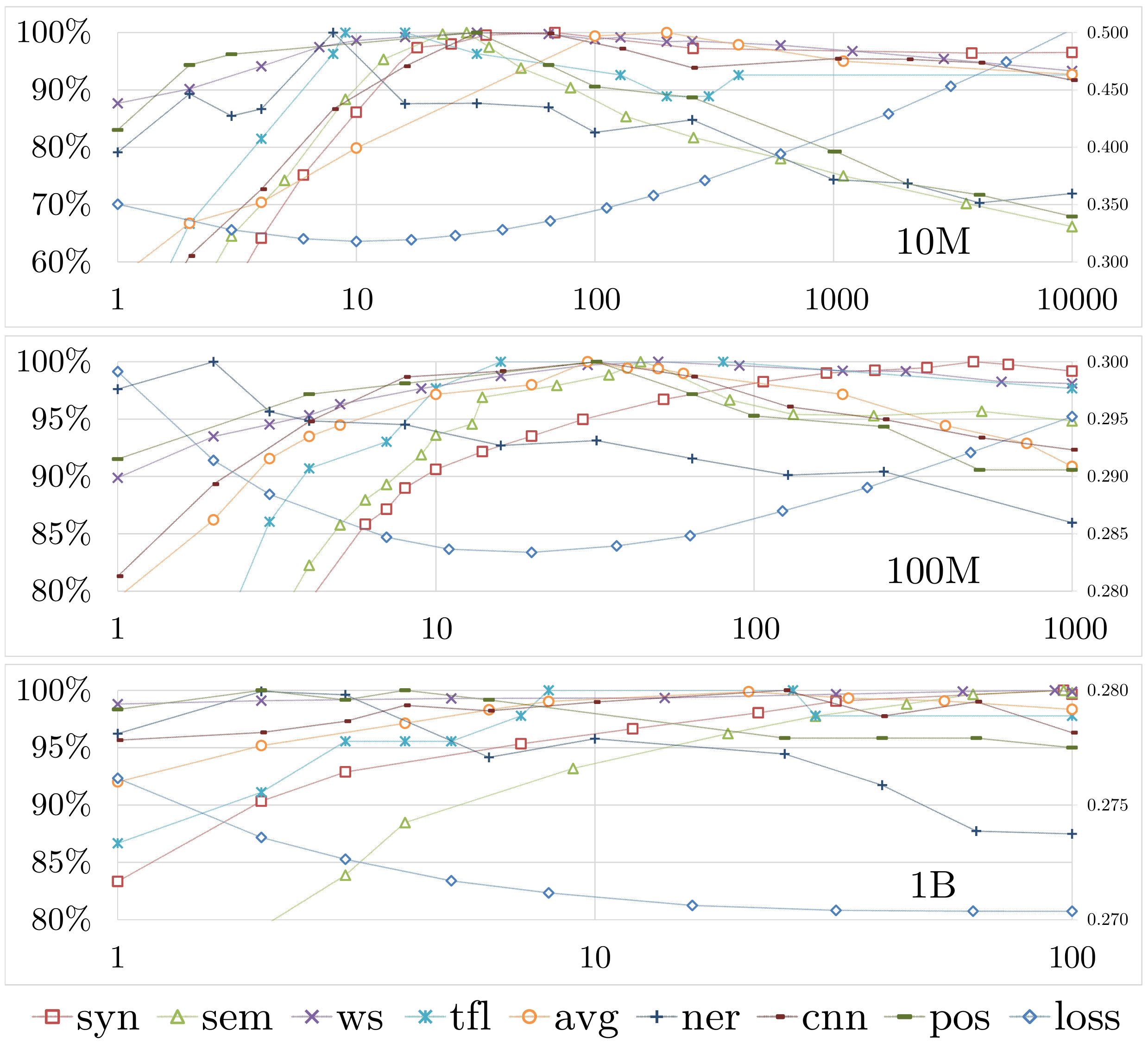}
	\caption{Task $PGR$s for embeddings trained with different numbers of iterations using the CBOW model on 10M-, 100M-, and 1B-token subsets of the W\&N corpus.}
	\label{fig:iter}
\end{figure}

In this experiment, we use 95\% of the corpus as the training set, and the remaining 5\% is used as the validation set. We report the loss on the validation set in addition to the eight evaluation tasks.
Figure~\ref{fig:iter} shows three examples of the training procedure. The embeddings are trained with the CBOW model on various subsets of the W\&N corpus.

From these cases, we find that the loss on the validation set is inconsistent with the performance in the real tasks. 
On the 100M-token corpus, the loss on the validation set peaks at the 20th iteration, which means that the model overfits the training data after the 20th iteration. However, the performance on almost all tasks continues to increase with further iterations. By contrast, on the 1B-token corpus, the model never overfits the training set, but the performances of the \textit{ner} and pos tasks decrease after several iterations. 

Despite the inconsistency between the validation loss and the task performance, it is worth noting that most tasks always exhibit similar peaks. The results demonstrate that we can use a simple task to verify whether the word embedding has peaked on other tasks. 
We consider the various combinations of the eight tasks, the seven models, and the three subsets of the W\&N corpus, thereby obtaining 168 tuples. Table~\ref{tab:iter_stop} shows the number of cases in which the word embedding model will win (achieve 95\% performance with respect to the peak performance on the task) when either the peak validation loss or some other task is used as the stopping criterion.
If we use the strategy that stop at the peak of the loss on the validation set, the resulting trained model will win in 89 cases. If we use the strategy that stops at the peak of the \textit{tfl} task (the simplest task in our experiments), it will win in 117 cases.

\begin{table}[ht]
	\centering
	\begin{tabular}{c|cccccccc}
		\toprule
		\textbf{loss} & \textbf{syn} & \textbf{sem} & \textbf{ws} & \textbf{tfl} & \textbf{avg} & \textbf{ner} & \textbf{cnn} & \textbf{pos}\tabularnewline
		\midrule
		89    & 105 & 111 & 103 & 117 & 104 & 91 & 103 & 101\tabularnewline
		\bottomrule
	\end{tabular}
	\caption{The number of cases in which the strategy wins if iteration is stopped when the specified task peaks.}
	\label{tab:iter_stop}
\end{table}

When training a word embedding for a specific task, using the development set for that task to determine when to stop iteration is the optimal choice because this approach will yield the results that are the most consistent with the task performance. However, in certain situations, testing the development set performance is time consuming. For example, the \textit{cnn} and \textit{pos} tasks require take dozens of minutes for a performance check, whereas the \textit{tfl} task requires only a few seconds. Therefore, this strategy provides a good approximation of the performance peak and is useful when testing the task performance would be excessively time consuming.

It should also be noted that multiple iterations are necessary. The performance increases by a large margin when we iterate more than once, regardless of the task and the corpus. Thus, the early version of the \texttt{word2vec} toolkit, which only scans the corpus once, may lose some performance as a result.

For the question of \emph{how many iterations should be applied to obtain a sufficient word embedding while avoiding overfitting?}, we argue that iterating the model until it peaks on some simple task will yield sufficient embeddings for most tasks.
To train a more suitable word embeddings for specific tasks, we can use the development set performance for that task to decide when to stop iterating.

\subsubsection{Dimensionality of the Embedding}
\label{sec:expr_dim}
\begin{figure}[th]
	\centering
	\includegraphics[width=0.47\textwidth]{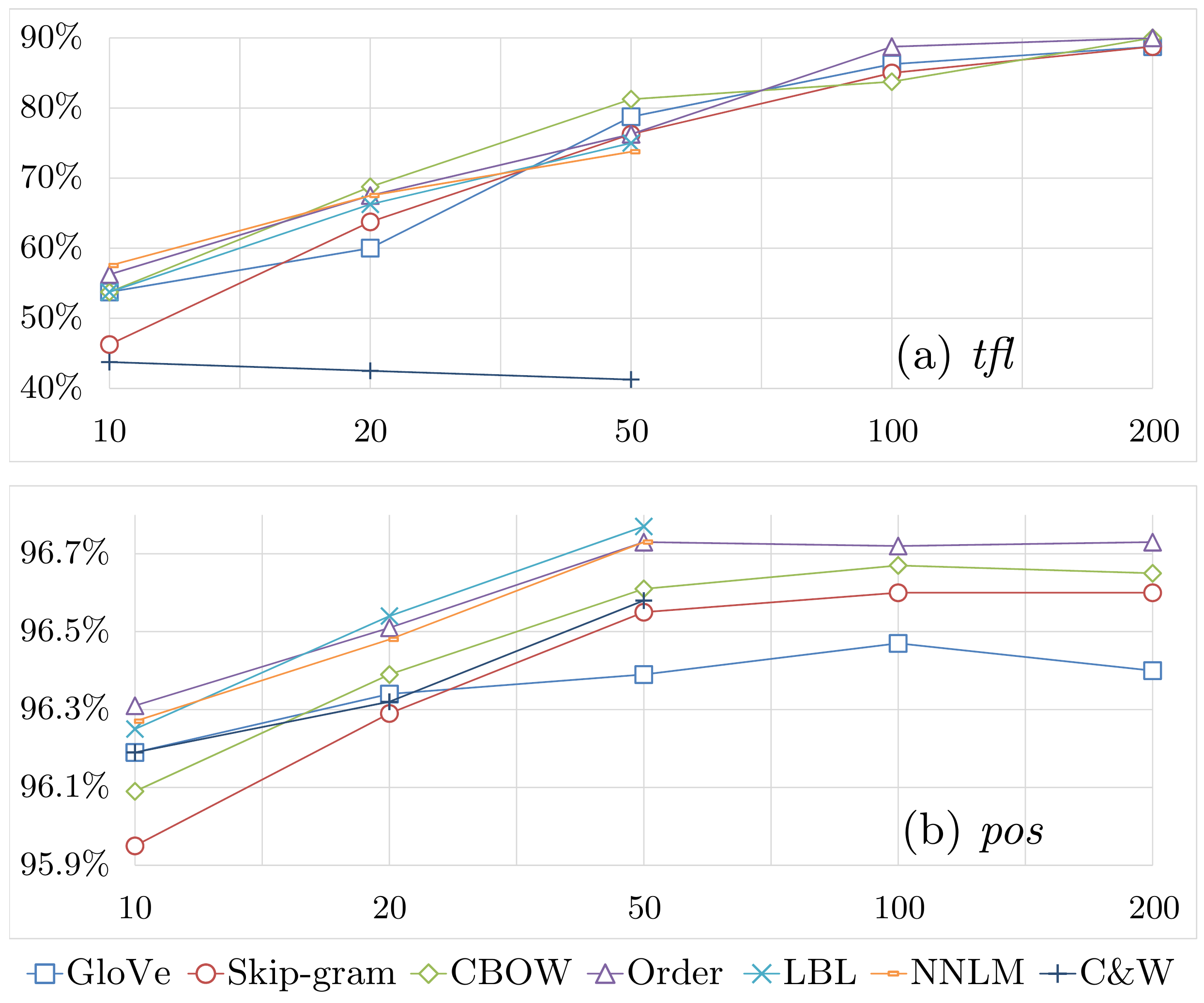}
	\caption{Task performances for embeddings of different dimensions trained using the various models on the 1B-token subset of the W\&N corpus.}
	\label{fig:dim}
\end{figure}

To investigate the influence of the dimensionality on word embedding training, we compare models of different dimensions in the eight evaluation tasks. Interestingly, the results show that the tasks that involve analyzing the embedding's semantic properties all behave in a similar manner. Figure~\ref{fig:dim}(a) shows the performance for the \textit{tfl} task as an example. Moreover, the tasks in which the embedding is used as a feature or for initialization also behave in a similar manner. Figure~\ref{fig:dim}(b) shows the performance for the \textit{pos} task as an example.

\emph{Which dimensionality should we choose to obtain a good enough embedding?}
We find that for the semantic property tasks, larger dimensions will lead to better performance (except for the C\&W model, as explained in Section~\ref{sec:expr_model}). However, for the NLP tasks, a dimensionality of 50 is typically sufficient.

\section{Related Work}
Regarding model comparison, we find that the most similar work to ours is Turian et al.~\cite{turian:2010}. In their work, the HLBL and C\&W models were compared for the NER and chunking tasks. They trained word embeddings on a relatively small corpus containing 63M tokens and concluded that the two embedding methods resulted in similar improvements to these tasks. 
Baroni et al.~\cite{baroni:2014don} compared count models (models based on a word-context co-occurrence matrix) and a predict model (the neural-network-based word embedding model; in their experiments, they used the CBOW model) on several tasks involving the evaluation of the semantic properties of words. They claimed that the predict model achieved significantly better performance than did the count model in almost all of the semantic tasks. However, 
Milajevs et al.~\cite{milajevs:2014} proposed that the co-occurrence matrix approaches should be more efficient than predict models for semantic compositional tasks.
Furthermore, Levy and Goldberg \cite{levy:2014} showed that the Skip-gram model with negative sampling implicitly factorizes a word-context PMI matrix, and their experimental results showed that the matrix approaches are competitive with the Skip-gram model for semantic similarity tasks.
Pennington et al.~\cite{Pennington:2014} proposed the matrix-based model GloVe and claimed that this model outperformed the CBOW and Skip-gram models. This result is contrary to our findings. We believe that the difference can be predominantly attributed to the different iteration strategies. In their experiments, they used the CBOW and Skip-gram models with a single iteration, whereas 25 iterations were performed for the GloVe model. Our experiment shows that performing multiple iterations can improve performance by a large margin. 

Several works have demonstrated the influence of corpus size and domain when comparing corpora. Most of them only discussed the corpus size. These works are not limited to word embedding; certain studies have also used cluster-based word representations.
Mikolov et al.~\cite{Mikolov:2013ICLR} observed that larger training corpora can lead to better performance of the CBOW model on analogy tasks (\textit{syn} and \textit{sem}). Pennington et al.~\cite{Pennington:2014} found that a larger corpus size is beneficial only for the \textit{syn} task and not the \textit{sem} task when training using the GloVe model.
Stenetorp et al.~\cite{stenetorp:2012size} found that training cluster-based word representations on in-domain corpora can improve performance in a biomedical NER task compared with training on newswire-domain corpora.

\section{Conclusion}
We fairly compared various word embedding models on various tasks and analyzed the critical components of word embedding training: the model, the corpus, and the training parameters. Although we did not find any specific settings for training a word embedding that can yield the best performance on all tasks (and indeed, such settings may not exist), we provided several guidelines for training word embeddings.

%
\bibliographystyle{abbrv}
\bibliography{compare}  
%
%

\end{document}